\documentclass[final]{cvpr}

\usepackage{times}
\usepackage{epsfig}
\usepackage{graphicx}
\usepackage{amsmath}
\usepackage{amssymb}

\usepackage{amsthm}
\usepackage{amsfonts} 
\usepackage{mathabx}
\usepackage{bm}
\usepackage{caption}
\usepackage{subcaption}
\usepackage{color}
\usepackage{multirow}
\usepackage{stackengine}
\usepackage{algorithm}
\usepackage[noend]{algpseudocode}
\usepackage{multirow}
\usepackage{soul}
\usepackage{tablefootnote}
\usepackage{makecell}

\def \W{\bm{\mathcal{W}}}
\def \T{\bm{\mathcal{T}}}

\def\Z{\bm{\mathcal{Z}}}

\def\U{\bm{\mathcal{U}}}

\DeclareMathOperator*{\argmin}{argmin}

\newcommand\blfootnote[1]{%
  \begingroup
  \renewcommand\thefootnote{}\footnote{#1}%
  \addtocounter{footnote}{-1}%
  \endgroup
}

\usepackage[pagebackref=true,breaklinks=true,colorlinks,bookmarks=false]{hyperref}

\def\cvprPaperID{5838} % *** Enter the CVPR Paper ID here
\def\confYear{CVPR 2021}

\begin{document}

%%%%%%%%% TITLE
\title{Towards Efficient Tensor Decomposition-Based DNN Model Compression with Optimization Framework }

\author{Miao Yin$^1$, Yang Sui$^1$, Siyu Liao$^{2\dagger}$ and Bo Yuan$^1$\\
$^1$Department of ECE, Rutgers University, $^2$Amazon\\
{\tt\small \{miao.yin, yang.sui\}@rutgers.edu, liasiyu@amazon.com, bo.yuan@soe.rutgers.edu}
% \and
}

\maketitle
% \pagestyle{empty}  % no page number for the second and the later pages
% \thispagestyle{empty} % no page number for the first page

%%%%%%%%% ABSTRACT
\begin{abstract}
Advanced tensor decomposition, such as tensor train (TT) and tensor ring (TR), has been widely studied for deep neural network (DNN) model compression, especially for recurrent neural networks (RNNs). However, compressing convolutional neural networks (CNNs) using TT/TR always suffers significant accuracy loss. In this paper, we propose a systematic framework for tensor decomposition-based model compression using Alternating Direction Method of Multipliers (ADMM). By formulating TT decomposition-based model compression to an optimization problem with constraints on tensor ranks, we leverage ADMM technique to systemically solve this optimization problem in an iterative way. During this procedure, the entire DNN model is trained in the original structure instead of TT format, but gradually enjoys the desired low tensor rank characteristics. We then decompose this uncompressed model to TT format, and fine-tune it to finally obtain a high-accuracy TT-format DNN model. Our framework is very general, and it works for both CNNs and RNNs, and can be easily modified to fit other tensor decomposition approaches. We evaluate our proposed framework on different DNN models for image classification and video recognition tasks. Experimental results show that our ADMM-based TT-format models demonstrate very high compression performance with high accuracy. Notably, on CIFAR-100, with 2.3$\times$ and 2.4$\times$ compression ratios, our models have 1.96\% and 2.21\% higher top-1 accuracy than the original ResNet-20 and ResNet-32, respectively. For compressing ResNet-18 on ImageNet, our model achieves 2.47$\times$ FLOPs reduction without accuracy loss. \blfootnote{$^\dagger$This work was done when the author was with Rutgers University.}

% Our evaluation on different DNN models for image classification and video recognition tasks show the excellent performance of this framework. On CIFAR-10, for ResNet-20 and ResNet-32 compression, with the same or even higher compression ratios, our ADMM-based TT-format models achieve $0.67\%\sim2.53\%$ and $1.36\%\sim1.43\%$ top-1 accuracy improvement, respectively, over the state-of-the-art tensor decomposed models. On CIFAR-100, not just outperforming the state-of-the-art works, our ADMM-based TT-format models even have 1.96\% and 2.21\% higher top-1 accuracy than the original ResNet-20 and ResNet-32 models, respectively. For compressing ResNet-18 on ImageNet, our approach achieves 2.0$\times$ compression ratio and 2.23$\times$ speedup with only XXXX accuracy loss, which outperforms several state-of-the-art model compression works. Besides, in video classification tasks, for compressing LSTM models on UCF11 and HMDB-51 datasets, our ADMM-based TT-format models also significantly outperform the state-of-the-art TT/TR compressed models with achieving higher accuracy and higher compression ratio simultaneously.

\end{abstract}

%%%%%%%%% BODY TEXT
\section{Introduction}

Deep Neural Network (DNNs) have already obtained widespread applications in many computer vision tasks, such as image classification \cite{he2016deep, krizhevsky2017imagenet},
%, simonyan2014very, szegedy2016rethinking}, 
video recognition \cite{feichtenhofer2016convolutional, carreira2017quo}, objective detection \cite{girshick2015fast, redmon2016you}, and image caption \cite{xu2015show, donahue2015long}. Despite these unprecedented success and popularity, executing DNNs on the edge devices is still very challenging. For most embedded and Internet-of-Things (IoT) systems, the sizes of many state-of-the-art DNN models are too large, thereby causing high storage and computational demands and severely hindering the practical deployment of DNNs. To mitigate this problem, to date many \textit{model compression} approaches, such as pruning \cite{han2015deep, han2015learning, luo2017thinet, zhang2018systematic} and quantization \cite{han2015learning, xu2018deep, rastegari2016xnor}, have been proposed to reduce the sizes of DNN models with limited impact on accuracy.

\textbf{Tensor Decomposition for Model Compression.} Recently, \textit{tensor decomposition}, as a mathematical tool that explores the low tensor rank characteristics of the large-scale tensor data, have become a very attractive DNN model compression technique. Different from other model compression methods, tensor decomposition, uniquely, can provide ultra-high compression ratio, especially for recurrent neural network (RNN) models. As reported in \cite{yang2017tensor,pan2019compressing}, the advanced tensor decomposition approaches, such as tensor train (TT) and tensor ring (TR), can bring more than 1,000$\times$ parameter reduction to the input-to-hidden layers of RNN models, and meanwhile the corresponding classification accuracy in the video recognition task can be even significantly improved. Motivated by such strong compression performance, many prior research works have been conducted on tensor decomposition-based DNN models \cite{garipov2016ultimate,novikov2015tensorizing,wang2018wide}. In addition, to fully utilize the benefits provided by those models, several TT-format DNN hardware accelerators have been developed and implemented in different chip formats, such as digital CMOS ASIC \cite{deng2019tie}, memristor ASIC \cite{huang2017ltnn} and IoT board \cite{cheng2019deepeye}.

\textbf{Limitations of the State of the Art.} Despite its promising potentials, the performance of tensor decomposition is not satisfied enough as a mature model compression approach. Currently all the reported success of tensor decomposition are narrowly limited to compressing RNN models in video recognition tasks. For compressing convolutional neural network (CNN) in the image classification task, which are the most commonly used and representative setting for evaluating model compression performance, all the state-of-the-art tensor decomposition approaches, including TT and TR, suffer very significant accuracy loss. For instance, even the very recent progress \cite{li2020heuristic} using TR still has 1.0$\%$ accuracy loss when the compression ratio is only 2.7$\times$ for ResNet-32 model on CIFAR-10 dataset. For the larger compression ratio as 5.8$\times$,  the accuracy loss further increases to 1.9\% .

\textbf{Why Limited Performance?} The above limitation of tensor decomposition is mainly due to the unique challenges involved in training the tensor decomposed DNN models. In general, there are two ways to use tensor decomposition to obtain a compressed model: 1) Train from scratch in the decomposed format; and 2) Decompose a pre-trained uncompressed model and then retrain. In the former case, when the required tensor decomposition-based, e.g. TT-format model, is directly trained from scratch, because the structure of the models are already pre-set to low tensor rank format before the training, the corresponding model capacity is typically limited as compared to the full-rank structure, thereby causing the training process being very sensitive to initialization and more challenging to achieve high accuracy. In the later scenario, though the pre-trained uncompressed model provides good initialization position, the straightforwardly decomposing full-rank uncompressed model into low tensor rank format causes inevitable and non-negligible approximation error, which is still very difficult to be recovered even after long-time re-training period. Besides, no matter which training strategy is adopted, tensor decomposition always brings linear increase in network depth, which implies training the tensor decomposition-format DNNs are typically more prone to gradient vanishing problem and hence being difficult to be trained well.

\textbf{Technical Preview and Contributions.} To overcome the current limitations of tensor decomposition and fully unlock its potentials for model compression, in this paper \ul{we propose a systematic framework for tensor decomposition-based model compression using alternating direction method of multipliers (ADMM)}. By formulating TT decomposition-based model compression to an optimization problem with constraints on tensor ranks, we leverage ADMM technique \cite{boyd2011distributed} to systemically solve this optimization problem in an iterative way. During this procedure the entire DNN model is trained in the original structure instead of TT format, but gradually enjoys the desired low tensor rank characteristics. We then decompose this uncompressed model to TT format, and fine-tune it to finally obtain a high-accuracy TT-format DNN model. In overall, the contributions of this paper are summarized as follows:
\begin{itemize}
    \item We propose a systematic framework to formulate and solve the tensor decomposition-based model compression problem. With formulating this problem to a constrained non-convex optimization problem, our framework gradually restricts the DNN model to the target tensor ranks without explicitly training on the TT format, thereby maintaining the model capacity as well as avoiding huge approximation error and increased network depth.
    \item We propose to use ADMM to efficiently solve this reformulated optimization problem via separately solving two sub-problems: one is to directly optimize the loss function with a regularization of the DNN by stochastic gradient descent, and the other is to use the introduced projection to constraint the tensor ranks analytically.
    \item We evaluate our proposed framework on different DNN models for image classification and video recognition tasks. Experimental results show that our ADMM-based TT-format models demonstrate very high compression performance with high accuracy. Notably, on CIFAR-100, with 2.3$\times$ and 2.4$\times$ compression ratios, our models have 1.96\% and 2.21\% higher top-1 accuracy than the original ResNet-20 and ResNet-32, respectively. For compressing ResNet-18 on ImageNet, our model achieves 2.47$\times$ FLOPs reduction with no accuracy loss.
    
    %  In particular, for compressing ResNet-18 on ImageNet, which is challenging for TT/TR-based compression, our ADMM-based TT-format model does not only significantly outperform existing TT/TR-based works, but also demonstrates better performance than other compression methods including pruning and matrix SVD. Out model achieves 2.0$\times$ compression ratio and 2.23$\times$ speedup with only XXXX accuracy loss. 

    % On CIFAR-10, for ResNet-20 and ResNet-32 compression, with the same or even higher compression ratios, our ADMM-based TT-format models achieve $0.67\%\sim2.53\%$ and $1.36\%\sim1.43\%$ top-1 accuracy improvement, respectively, over the state-of-the-art tensor decomposed models. On CIFAR-100, not just outperforming the state-of-the-art works, our ADMM-based TT-format models even have 1.96\% and 2.21\% higher top-1 accuracy than the original ResNet-20 and ResNet-32 models, respectively. For compressing ResNet-18 on ImageNet, our approach achieves 2.0$\times$ compression ratio and 2.23$\times$ speedup with only XXXX accuracy loss, which outperforms several state-of-the-art model compression works. Besides, in video classification tasks, for compressing LSTM models on UCF11 and HMDB-51 datasets, our ADMM-based TT-format models also significantly outperform the state-of-the-art TT/TR compressed models with achieving higher accuracy and higher compression ratio simultaneously.

\end{itemize}

\section{Related Work on DNN Model Compression}

\textbf{Sparsification.} Sparsification is the most popular DNN compression approach. Different levels of network structure can be sparse, such as weight \cite{han2015deep,han2015learning}, filter \cite{luo2017thinet,he2018soft} and channel \cite{he2017channel,zhuang2018discrimination}. To obtain the sparsity, a DNN model can be either pruned \cite{han2015deep,luo2017thinet} or trained with sparsity-aware regularization \cite{liu2015sparse,zhou2016less}. Also, the introduced sparsity can be either structured or unstructured. Unstructured sparse models \cite{han2015deep, zhang2018systematic} enjoy high accuracy and compression ratio, but brings irregular memory access and imbalanced workload problems \cite{han2016eie} to the underlying hardware platform. Structured sparse models \cite{wen2016learning} are more hardware friendly; however, their compression ratio and accuracy are typically inferior to the unstructured counterparts.

\textbf{Quantization.} Quantization \cite{rastegari2016xnor,han2015deep,xu2018deep} is another widely adopt model compression approach. By reducing the number of bits for weight representation, quantization enables immediate reduction in DNN model size. The most aggressive quantization scheme brings binary networks \cite{rastegari2016xnor,courbariaux2015binaryconnect}, which only use 1-bit weight parameters. Quantization is inherently hardware friendly, and have become a standard adopted model compression method for most DNN hardware accelerators \cite{han2016eie,chen2016eyeriss,jouppi2017datacenter}. However, quantization is limited by the maximum compression ratio that can be offered (up to 32$\times$).

\textbf{Tensor Decomposition.} Rooted in tensor theory, tensor decomposition approach factorizes weight tensors into smaller tensors to reduce model sizes. In \cite{jaderberg2014speeding}, matrix-oriented singular value decomposition (SVD), as the low-dimensional instance of tensor decomposition, is used to perform model compression. However, using this method, or other classical high-dimensional tensor decomposition methods, such as Tucker \cite{tucker1963implications} and CP decomposition \cite{harshman1970foundations}, causes significant accuracy loss ($>$ 0.5\%) with limited compression ratios \cite{kim2015compression,lebedev2014speeding,gusak2019automated,phan2020stable}. Starting from \cite{garipov2016ultimate}, advanced tensor decomposition approaches, such as tensor train (TT) and tensor ring (TR) decomposition, have become the more popular options. These methods have very attractive advantages -- the compression ratio can be very high (e.g. $> \textrm{1,000}\times$) because of their unique mathematical property. Such benefits have been demonstrated on RNN compression in video recognition tasks. As reported in \cite{yang2017tensor,pan2019compressing}, 17,560$\times$ to 34,203$\times$ compression ratios can be achieved by using TT or TR decomposition on the input-to-hidden layer of RNN models for video recognition. However, TT and TR approaches do not perform well on CNN models. For instance, even the very recent progress \cite{wang2018wide,li2020heuristic} still suffers 1.0\% accuracy loss with 2.7$\times$ compression ratio, or even 1.9\% accuracy loss with 5.8$\times$ compression ratio, both for ResNet-32 model on CIFAR-10 dataset. From the perspective of practical deployment, such non-negligible accuracy degradation severely hinders the widespread adoption of tensor decomposition for many CNN-involved model compression scenarios.

% demonstrated effective but with low compression ratio \cite{denton2014exploiting}.  
% Instead, tensor decomposition methods have successfully pushed the limit of low rank representation. 
% For example, there are works studying tensor train (TT) \cite{yang2017tensor}, tensor ring (TR) \cite{pan2019compressing}, and block term (BT) \cite{ye2018learning} decomposition. 
% These methods have achieved state-of-the-art results but they all focus on recurrent neural networks that mainly consists of weight matrices (XXX should specify the redundancies of RNNs).

% On the other hand, Tucker decomposition \cite{kim2015compression} has been applied to weight tensors in convolution layer and weight matrices in fully-connected layers, where they factorize pre-trained model and then perform finetuning to recover model performance. 
% In \cite{garipov2016ultimate}, TT method is applied to convolution layer by factorizing the corresponding matrix representation of the weight tensor. 
% Given that tensor decomposition are often trained under fixed rank setting, 
% \cite{li2020heuristic} applies a heuristic rank search algorithm over pre-trained model to form TR factorized CNN model.
% In summary, \cite{kim2015compression,garipov2016ultimate,li2020heuristic} are generalized to compress both convolution layer and fully-connected layer, but they all use the pre-trained models, which significantly increase the cost. 

\section{Background and Preliminaries}
% Considering TT networks are the first tensor networks and the most popular ones, in this work we implement the proposed framework based on TT networks instead of all tensor networks, thus in this section we only introduce the background of TT such as TT decomposition and TT layers.
\subsection{Notation}
$\bm{\mathcal{X}}\in\mathbb{R}^{n_1 \times n_2 \times \cdots \times n_d}$,  $\bm{X}\in\mathbb{R}^{n_1\times n_2}$, and $\bm{x}\in\mathbb{R}^{n_1}$ represent $d$-order tensor, matrix and vector, respectively. Also,  $\bm{\mathcal{X}}_{(i_1,\cdots,i_d)}$ and $\bm{X}_{(i,j)}$ denote the single entry of tensor $\bm{\mathcal{X}}$ and matrix $\bm{X}$, respectively. 
%, and $\bm{\mathcal{X}}_{(i_1,\cdots,i_k)}\in\mathbb{R}^{n_{k+1}\times\cdots\times n_{d}}$ denotes the specific sub-tensor. 
% Similarly,  represents the entry of matrix $\bm{X}$.

%\textbf{Tensor Contraction.} Tensor contraction is executed between two tensors with at least one matched dimension. For instance, given two tensors $\bm{\mathcal{A}}\in\mathbb{R}^{n_1\times n_2 \times l}$ and $\bm{\mathcal{B}}\in\mathbb{R}^{l\times m_1\times m_2}$, where the 3rd dimension of $\bm{\mathcal{A}}$ matches the 1st dimension of $\bm{\mathcal{B}}$ with length $l$, the tensor contraction result is a size- $n_1\times n_{2}\times m_1 \times m_2$ tensor as
%$$ 
%(\bm{\mathcal{A}}\times\bm{\mathcal{B}})_{(i_1, i_2, j_1, j_2)}=\sum_{k=1}^{l}\bm{\mathcal{A}}_{(i_1, i_2, k)}\bm{\mathcal{B}}_{(k, j_1, j_2)}.
%$$
\subsection{Tensor Train (TT) Decomposition} 
\begin{figure}[t]
% \vspace{-2mm}
    \centering
    \includegraphics[width=\linewidth]{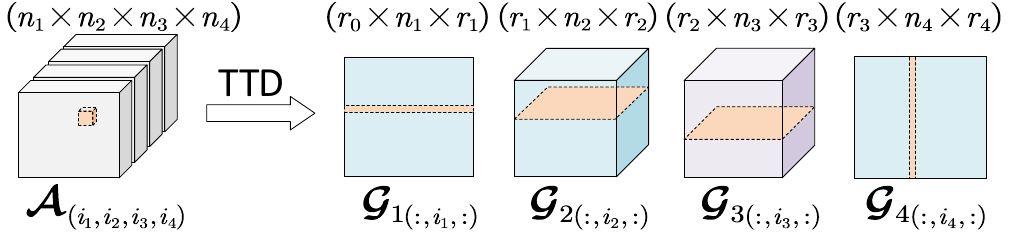}
    \vskip -3mm
    \caption{Illustration of Tensor Train Decomposition (TTD) for a 4-order tensor. $r_0$ and $r_4$ are always equal to 1.}
    \label{fig:ttd}
    \vspace{-4mm}
\end{figure}

Given a tensor $\bm{\mathcal{A}}\in\mathbb{R}^{n_1\times n_2\times\cdots\times n_d}$, it can be decomposed to a sort of 3-order tensors via Tensor Train Decomposition (TTD) as follows:\vskip -5mm
\begin{equation}
\begin{aligned}
    \bm{\mathcal{A}}_{(i_1,i_2,\cdots,i_d)}&={\bm{\mathcal{G}}_1}_{(:,i_1,:)}{\bm{\mathcal{G}}_2}_{(:,i_2,:)}\cdots{\bm{\mathcal{G}}_d}_{(:,i_d,:)}\\
    &~\begin{aligned}=\sum_{\alpha_0,\alpha_1\cdots\alpha_d}^{r_0,r_1,\cdots r_d}&{\bm{\mathcal{G}}_1}_{(\alpha_0,i_1,\alpha_1)}{\bm{\mathcal{G}}_2}_{(\alpha_1,i_2,\alpha_2)}\cdots\\
    &{\bm{\mathcal{G}}_d}_{(\alpha_{d-1},i_d,\alpha_d)},\end{aligned}
\end{aligned}
\label{eq1}
\vspace{-1mm}
\end{equation}
where $\bm{\mathcal{G}}_k\in\mathbb{R}^{r_{k-1}\times n_k\times r_k}$ are called \textit{TT-cores} for $k=1,2,\cdots,d$, and $\bm{r}=[r_0, r_1, \cdots, r_d], r_0=r_d=1$ are called \textit{TT-ranks}, which determine the storage complexity of TT-format tensor. An example is demonstrated in Figure \ref{fig:ttd}. 

% With target TT-ranks $\bm{r}^*=[r_0^*,r_1^*,\cdots,r_d^*]$, \textit{TT-SVD} \cite{oseledets2011tensor}, which iteratively applies matrix SVD and truncates the ranks on the reshaped matrices, can be used to obtain TT-cores that satisfy $r_1\le r_1^*,\cdots, r_{d-1}\le r_{d-1}^*, r_0=r_d=1$.

%\textbf{Mode-$(1,\cdots,i)$ Matricization.}  Given a tensor $\bm{\mathcal{A}}\in\mathbb{R}^{n_1\times n_2\times\cdots\times n_d}$, the mode-$(1,\cdots,i)$ matricization of $\bm{\mathcal{A}}$ is denoted as $\bm{\mathcal{A}}_{[i]}\in\mathbb{R}^{p_i\times q_i}$, where $p_i=\prod_{k=1}^{i}n_k, q_i=\prod_{k=i+1}^{d}n_k$. This operation can be obtained by using \texttt{reshape} function in MATLAB or Python environment, i.e. 
%$$
%\bm{\mathcal{A}}_{[i]}=\texttt{reshape}(\bm{\mathcal{A}}, [p_i, q_i]).
%$$
%The TT-ranks $r_1,\cdots,r_{d-1}$ of $\bm{\mathcal{A}}$ are also the matrix ranks of $\bm{\mathcal{A}}_{[1]},\cdots,\bm{\mathcal{A}}_{[d-1]}$, which can be calculated by matrix SVD, as described by \cite{oseledets2011tensor} in detail.

\subsection{Tensor Train (TT)-format DNN}

\textbf{TT Fully-Connected Layer.} Consider a simple fully-connected layer with weight matrix $\bm{W}\in\mathbb{R}^{M\times N}$ and input $\bm{x}\in\mathbb{R}^N$, where $M=\prod_{k=1}^{d}m_k$ and $N=\prod_{k=1}^{d}n_k$, the output $\bm{y}\in\mathbb{R}^{M}$ is obtained by $\bm{y}=\bm{W}\bm{x}$. In order to transform this standard layer to TT fully-connected (TT-FC) layer, we first tensorize the weight matrix $\bm{W}$ to a weight tensor $\bm{\mathcal{W}}\in\mathbb{R}^{(m_1\times n_1)\times\cdots\times(m_d\times n_d)}$ by reshaping and order transposing. Then $\bm{\mathcal{W}}$ can be decomposed to TT-format:\vskip -7mm
\begin{equation}
\bm{\mathcal{W}}_{((i_1,j_1),\cdots,(i_d,j_d))} = {\bm{\mathcal{G}}_1}_{(:,i_1,j_1,:)}\cdots{\bm{\mathcal{G}}_d}_{(:,i_d,j_d,:)}.
\label{eq2}
\vspace{-2mm}
\end{equation}
Here, each TT-core $\bm{\mathcal{G}}_k\in\mathbb{R}^{r_{k-1}\times m_k\times n_k\times r_k}$ is a 4-order tensor, which is one dimension more than the standard one since the output and input dimensions of $\bm{W}$ are divided separately. Hence, the forward progagation on the TT-FC layer can be expressed in tensor format as follows:\vskip -6mm
\begin{equation}
\begin{aligned}
\bm{\mathcal{Y}}_{(i_1,\cdots,i_d)}=\sum_{j_1,\cdots,j_d} {\bm{\mathcal{G}}_1}_{(:,i_1,j_1,:)}\cdots{\bm{\mathcal{G}}_d}_{(:,i_d,j_d,:)}\bm{\mathcal{X}}_{(j_1,\cdots,j_d)},
\end{aligned}
\label{eq3}
\vspace{-0.5mm}
\end{equation}
where $\bm{\mathcal{X}}\in\mathbb{R}^{m_1\times\cdots\times m_d}$ and $\bm{\mathcal{Y}}\in\mathbb{R}^{n_1\times\cdots\times n_d}$ are the tensorized input and output corresponding to $\bm{x}$ and $\bm{y}$, respectively. The details about TT-FC layer is introduced in \cite{novikov2015tensorizing}.

\begin{figure*}[ht!]
    \centering
    \includegraphics[width=\linewidth]{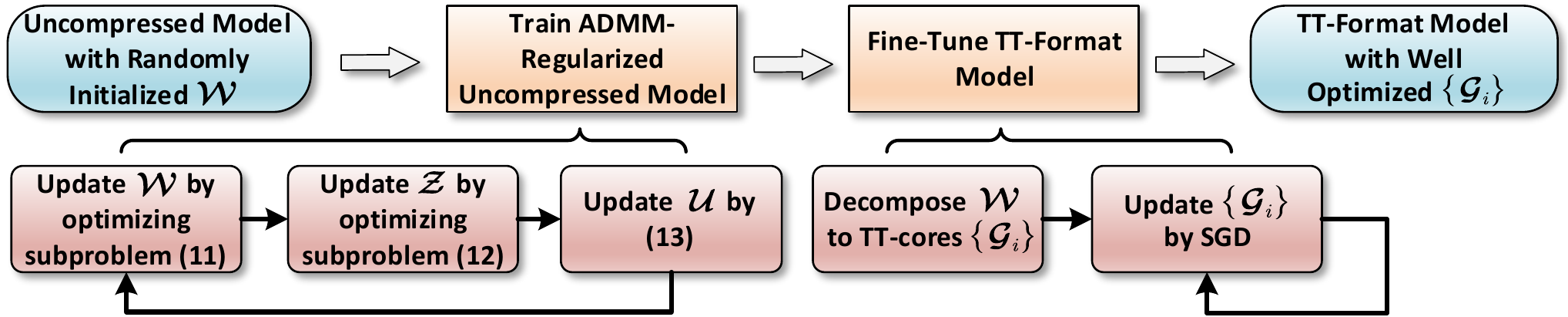}
    \vspace{-7mm}
    \caption{Procedure of the proposed compression framework using ADMM for a TT-format DNN model.}
    \label{fig:admm}
    \vspace{-4mm}
\end{figure*}

\textbf{TT Convolutional Layer.} For a conventional convolutional layer, its forward computation is to perform convolution between a 3-order input tensor $\widetilde{\bm{\mathcal{X}}}\in\mathbb{R}^{W\times H\times N}$ and a 4-order weight tensor $\widetilde{\bm{\mathcal{W}}}\in\mathbb{R}^{K\times K\times M\times N}$ to produce the 3-order output tensor $\widetilde{\bm{\mathcal{Y}}}\in\mathbb{R}^{(W-K+1)\times (H-K+1)\times M}$ . In a TT convolutional (TT-CONV) layer, the input tensor $\widetilde{\bm{\mathcal{X}}}$ is reshaped to a tensor $\bm{\mathcal{X}}\in\mathbb{R}^{W\times H\times n_1\times\cdots\times n_d}$, while the weight tensor $\widetilde{\bm{\mathcal{W}}}$ is reshaped and transposed to a tensor $\bm{\mathcal{W}}\in\mathbb{R}^{(K\times K)\times(m_1\times n_1)\times\cdots\times(m_d\times n_d)}$ and then decomposed to TT-format:\vskip -7mm
\begin{equation}
\begin{aligned}
\bm{\mathcal{W}}_{((k_1,k_2),(i_1,j_1),\cdots,(i_d,j_d))}=&{\bm{\mathcal{G}}_0}_{(k_1,k_2)}{\bm{\mathcal{G}}_1}_{(:,i_1,j_1,:)}\cdots\\
&{\bm{\mathcal{G}}_d}_{(:,i_d,j_d,:)},
\end{aligned}
\label{eq4}
\vspace{-2mm}
\end{equation}
where $M=\prod_{k=1}^{d}m_k$ and $N=\prod_{k=1}^{d}n_k$. Similar with TT-FC layer, here $\bm{\mathcal{G}}_k\in\mathbb{R}^{r_{k-1}\times m_k\times n_k\times r_k}$ is a 4-order tensor except $\bm{\mathcal{G}}_0\in\mathbb{R}^{K\times K}$. Then the new output tensor $\bm{\mathcal{Y}}\in\mathbb{R}^{(W-K+1)\times(H-K+1)\times m_1\times\cdots\times m_d}$ is obtained by\vskip -7mm
\begin{equation}
\begin{aligned}
\bm{\mathcal{Y}}_{(w,h,i_1,\cdots,i_d)}&=\\\sum_{k_1=1}^{K}\sum_{k_2=1}^{K}&\sum_{j_1,\cdots,j_d}\bm{\mathcal{X}}_{(k_1+w-1,k_2+h-1,j_1,\cdots,j_d)}\\&{\bm{\mathcal{G}}_0}_{(k_1,k_2)}{\bm{\mathcal{G}}_1}_{(:,i_1,j_1,:)}\cdots{\bm{\mathcal{G}}_d}_{(:,i_d,j_d,:)}.
\end{aligned}
\label{eq5}
\vspace{-2mm}
\end{equation}
The detailed description of TT-CONV layer is in \cite{garipov2016ultimate}.

\textbf{Training on TT-format DNN.} As TT-FC layer, TT-CONV layer and the corresponding forward propagation schemes are formulated, standard stochastic gradient descent (SGD) algorithm can be used to update the TT-cores with the rank set $\bm{r}$, which determines the target compression ratio. The initialization of the TT-cores can be either randomly set or obtained from directly TT-decomposing a pre-trained uncompressed model.

\section{Systematic Compression Framework}

\textbf{Analysis on Existing TT-format DNN Training.} As mentioned in the last paragraph, currently a TT-format DNN is either 1) trained from with randomly initialized tensor cores; or 2) trained from a direct decomposition of pre-trained model. For the first strategy, it does not utilize any information related to the high-accuracy uncompressed model; while other model compression methods, e.g. pruning and knowledge distillation, have shown that proper utilization of the pre-trained models are very critical for DNN compression. For the second strategy, though the knowledge of the pre-trained model is indeed utilized, because the pre-trained model generally lacks low TT-rank property, after direct low-rank tensor decomposition the approximation error is too significant to be properly recovered even using long-time re-training. Such inherent limitations of the existing training strategies, consequently, cause significant accuracy loss for the compressed TT-format DNN models.

\textbf{Our Key Idea.} We believe the key to overcome these limitations is to maximally retain the knowledge contained in the uncompressed model, or in other words, minimize the approximation error after tensor decomposition with given target tensor ranks. To achieve that, we propose to formulate an optimization problem to minimize the loss function of the uncompressed model with low tensor rank constraints. With proper advanced optimization technique (e.g. ADMM)-regularized training procedure, the uncompressed DNN models can gradually exhibit low tensor rank properties. After the ADMM-regularized training phase, the approximation error brought by the explicit low-rank tensor decomposition becomes negligible, and can be easily recovered by the SGD-based fine-tuning. Figure \ref{fig:admm} shows the main steps of our proposed overall framework.

\subsection{Problem Formulation}

As mentioned above, the first phase of our framework is to gradually impose low tensor rank characteristics onto a high-accuracy uncompressed DNN model. Mathematically, this goal can be formulated as a optimization problem to minimize the loss function of the object model with constraints on TT-ranks of each layer (convolutional or fully-connected):\vskip -5mm
\begin{equation}
\begin{aligned}
\min_{\bm{\mathcal{W}}}~~&\ell(\bm{\mathcal{W}}),\\\vspace{-1mm}
\textrm{s.t.}~~~&\mathrm{rank}(\bm{\mathcal{W}})\le \bm{r}^*, 
\end{aligned}
\label{eqn:obj}
\vspace{-1.5mm}
\end{equation}
where $\ell$ is the loss function of the DNN , $\mathrm{rank}(\cdot)$ is a function that returns the TT-ranks $\bm{r}=[r_{0},\cdots,r_{d}]$ of the weight tensor cores, and $\bm{r}^*=[r_0^*,\cdots,r_d^*]$ are the desired TT-ranks for the layer. To simplify the notation, here $\bm{r}\le \bm{r}^*$ means $r_{i}\le r_{i}^*, i=0,\cdots,d$, for each $r_{i}$ in $\bm{r}$.

% Unlike the prior tensor decomposition-based DNN training, here we introduces a hard rank constraint i.e. $\mathrm{rank}(\bm{\mathcal{W}})\le\bm{r}^*$ of dense networks to the objective. By optimizing the proposed problem, the dense layers are restricted to the target TT-ranks, while the loss function of the original networks is also minimized. 

\subsection{Optimization Using ADMM}

Obviously, solving the problem (\ref{eqn:obj}) is generally difficult via using normal optimization algorithms since $\mathrm{rank}(\cdot)$ is non-differentiable. To overcome this challenge, we first rewrite it as\vskip -3mm
\begin{equation}
\begin{aligned}
\min_{\bm{\mathcal{W}}}~~&\ell(\bm{\mathcal{W}}),\\
\textrm{s.t.}~~~&\W\in\mathcal{S}, 
\end{aligned}
\label{eqn:obj_set}
\vspace{-1mm}
\end{equation}
where $\mathcal{S}=\{\W~|~\mathrm{rank}(\W)\le \bm{r}^*\}$. Hence, the objective form (\ref{eqn:obj_set}) is a classic non-convex optimization problem with constraints, which can be properly solved by ADMM \cite{boyd2011distributed}. Specifically, we can first introduce an auxiliary variable $\bm{\mathcal{Z}}$ and an indicator function $g(\cdot)$ of $\mathcal{S}$, i.e.\vskip -2mm
\begin{equation}
g(\bm{\mathcal{W}})= \begin{cases}
0 & \W \in \mathcal{S},\\ 
+\infty & \textrm{otherwise}.
\end{cases}
\vspace{-0.5mm}
\end{equation}
And then the problem (\ref{eqn:obj_set}) is equivalent to the following form:\vskip -5mm
\begin{equation}
\begin{aligned}
\min_{\bm{\mathcal{W}},\bm{\mathcal{Z}}}~~& 
\ell \left ( \bm{\mathcal{W}} \right ) + g(\bm{\mathcal{Z}}), \\
\textrm{s.t.}~~~ & \bm{\mathcal{W}} = \bm{\mathcal{Z}}.
\end{aligned}
\label{eqn:obj_w_z}
\vspace{-1mm}
\end{equation}
To ensure convergence without assumptions like strict convexity or finiteness of $\ell$, instead of Lagrangian, the corresponding augmented Lagrangian in the scaled dual form of the above problem is given by\vskip -6mm
\begin{equation}
\begin{aligned}
\mathcal{L}_{\rho}(\W, \Z, \U) = &\ell(\W) + g(\Z)\\
&+ \frac{\rho}{2} \left \| \W - \Z + \U\right \|_{F}^{2} + \frac{\rho}{2}\|\U\|_F^2,
\end{aligned}
\label{eqn:lagran}
\vspace{-1mm}
\end{equation}
where $\U$ is the dual multiplier, and $\rho>0$ is the penalty parameter. Thus, the iterative ADMM scheme can be explicitly performed as \vskip -7mm
\begin{align}
\W^{t+1}& = \argmin_{\W}~~\mathcal{L}_{\rho}\left ( \W, \Z^{t}, \U^{t} \right ), \label{eqn:sub_w} \\
\Z^{t+1}& = \argmin_{\Z}~~\mathcal{L}_{\rho}\left ( \W^{t+1}, \Z, \U^{t} \right ), \label{eqn:sub_z} \\
\U^{t+1}& = \U^{t} + \W^{t+1} - \Z^{t+1}, \label{eqn:sub_u}
% \vspace{-9mm}
\end{align}
\vskip -3mm \noindent where $t$ is the iterative step. Now, the original problem (\ref{eqn:obj_w_z}) is separated to two subproblems (\ref{eqn:sub_w}) and (\ref{eqn:sub_z}), which can be solved individually. Next, we introduce the detailed solution of each subproblem.

\textbf{$\W$-subproblem.}  The $\W$-subproblem (\ref{eqn:sub_w}) can be reformulated as follows: \vskip -3mm
\begin{equation}
\begin{aligned}
\min_{\W} &\quad \ell(\W) + \frac{\rho}{2} \left \| \W - \Z^{t} + \U^{t}\right \|_{F}^{2},
\end{aligned}
\label{eqn:min_w}
\vspace{-1mm}
\end{equation}
where the first term is the loss function, e.g. cross-entropy loss in classification tasks, of the DNN model, and the second term is the $L_2$ regularization. This subproblem can be directly solved by SGD since both these two terms are differentiable. Correspondingly, the partial derivative of (\ref{eqn:min_w}) with respect to $\W$ is calculated as \vskip -5mm
\begin{equation}
\begin{aligned}
\frac{\partial \mathcal{L}_{\rho}(\W, \Z^t, \U^t)}{\partial \W} = \frac{\partial \ell(\W)}{\partial \W} + \rho(\W-\Z^t+\U^t).
\end{aligned}
\label{eqn:w_loss}
\vspace{-1mm}
\end{equation}
And hence $\W$ can be updated by \vskip -2.5mm
\begin{equation}
\begin{aligned}
\W^{t+1} = \W^{t} - \eta\frac{\partial \mathcal{L}_{\rho}(\W, \Z^t, \U^t)}{\partial \W},
\end{aligned}
\label{eqn:w_update}
\vspace{-1mm}
\end{equation}
where $\eta$ is the learning rate.

\textbf{$\Z$-subproblem.} To solve $\Z$-subproblem (\ref{eqn:sub_z}), we first explicitly formulate it as follows: \vskip -2mm
\begin{equation}
\begin{aligned}
\min_{\Z} ~~g(\Z) + \frac{\rho}{2} \left \| \W^{t+1} - \Z + \U^{t}\right \|_{F}^{2},
\end{aligned}
\label{eqn:min_z}
\vspace{-1mm}
\end{equation}
where the indicator function $g(\cdot)$ of the non-convex set $\mathcal{S}$ is non-differentiable. Then, according to \cite{boyd2011distributed}, in this format  updating $\Z$ can be performed as:\vskip -3mm
\begin{equation}
\begin{aligned}
\Z^{t+1} = \bm{\Pi}_{\mathcal{S}}(\W^{t+1} + \U^t),
\end{aligned}
\label{eqn:z_update}
\end{equation}
\vskip -1mm\noindent where $\bm{\Pi}_{\mathcal{S}}(\cdot)$ is the projection of singular values onto $\mathcal{S}$, by which the TT-ranks of $(\W^{t+1}+\U^{t})$ are truncated to target ranks $\bm{r}^*$. Algorithm \ref{alg:ttsvd} describes the specific procedure of this projection in the TT-format scenario. 

\begin{algorithm}[ht] 
\caption{TT-SVD-based Projection for Solving (\ref{eqn:min_z})} 
\label{alg:ttsvd} 
\begin{algorithmic}[1] 
\Require 
$d$-order tensor $\bm{\mathcal{A}}\in\mathbb{R}^{n_1\times\cdots\times n_d}$, target TT-ranks $\bm{r}^*$.
\Ensure 
%$\hat{\bm{\mathcal{A}}} \in \bm{\mathcal{C}}(\bm{\mathcal{A}})=\{\bm{\mathcal{A}}~|~\mathrm{rank}(\bm{\mathcal{A}})\le\bm{r}^*\}$.
$\hat{\bm{\mathcal{A}}}=\bm{\Pi}_{\mathcal{S}}(\bm{\mathcal{A}})$.

\State Temporary tensor $\bm{\mathcal{T}}=\bm{\mathcal{A}}$;
\For{$k=1$ to $d-1$}
\State $\bm{\mathcal{T}}:=\texttt{reshape}(\bm{\mathcal{T}},[r_{k-1}^*n_k,-1])$;
\State Compute matrix SVD: $\bm{U}, \bm{S}, \bm{V}:=\texttt{SVD}(\bm{\mathcal{T}})$; 
\State $\bm{U}:=\bm{U}_{(1:r_k^*,:)}$;
\State $\bm{S}:=\bm{S}_{(1:r_k^*,1:r_k^*)}$;
\State $\bm{V}:=\bm{V}_{(:,1:r_k^*)}$;
\State $\bm{\mathcal{G}}_k:=\texttt{reshape}(\bm{U},[r_{k-1}^*,n_k,r_{k}^*])$;
\State $\bm{\mathcal{T}}:=\bm{S}\bm{V}^T$;
\EndFor
\State $\T:=\bm{\mathcal{G}}_1$;
\For{$k=1$ to $d-1$}
\State $\bm{T}_1:=\texttt{reshape}(\T,[-1,r_{k}^*])$;
\State $\bm{T}_2:=\texttt{reshape}(\bm{\mathcal{G}}_{k+1},[r_{k}^*,-1])$;
\State $\T:=\bm{T}_1\bm{T}_2$;
\EndFor
\State $\hat{\bm{\mathcal{A}}}=\texttt{reshape}(\T,[n_1,\cdots,n_d])$.

\end{algorithmic} 
\end{algorithm}

% As mentioned above, $\W$-minimization step and $\Z$-minimization step are updated one time separately before executing scaled dual variable update step. 
% Furthermore, we can also alternatively update above two steps iteratively $K$ times instead of one time before the scaled dual variable update step. in order to get the ideally optimal $\W$ of \ref{eq8}.

In each ADMM iteration, upon the update of $\W$ and $\Z$, the dual multiplier $\U$ is updated by (\ref{eqn:sub_u}). In overall,  to solve (\ref{eqn:obj_w_z}), the entire ADMM-regularized training procedure is performed in an iterative way until convergence or reaching the pre-set maximum iteration number. The overall procedure is summarized in Algorithm \ref{alg:admm}.

%After above two steps finished, $\W^{t+1}$ and $\Z^{t+1}$ are given to carry out $\U$-minimization step, which is updated literally by equation (\ref{eqn:sub_u}).

\begin{algorithm}[t] 
\caption{ADMM-Regularized Training Procedure}
\label{alg:admm} 
\begin{algorithmic}[1] 
\Require 
Weight tensor $\bm{\mathcal{W}}$, target TT-ranks $\bm{r}^*$, penalty parameter $\rho$, feasibility tolerance $\epsilon$, maximum iterations $T$.
\Ensure 
%$\hat{\bm{\mathcal{A}}} \in \bm{\mathcal{C}}(\bm{\mathcal{A}})=\{\bm{\mathcal{A}}~|~\mathrm{rank}(\bm{\mathcal{A}})\le\bm{r}^*\}$.
Optimized $\W$.

\State Randomly initialize $\W$; 
\State $\Z:=\W,~\U:= \bm{0}$;
%, \Z,\U\in \mathbb{R}^{n_1\times\cdots\times n_d}$
\While {$\|\W^{t} - \Z^{t}\| > \epsilon $ and $t \leq T$}
\State Updating $\W$ via (\ref{eqn:w_update});
\State Updating $\Z$ via (\ref{eqn:z_update}) (Algorithm \ref{alg:ttsvd});
\State Updating $\U$ via (\ref{eqn:sub_u});
\EndWhile
\State \textbf{end}

\end{algorithmic} 
\end{algorithm}

%\textbf{Stopping Criteria.}
%In the end, the entire training process ends up with the stopping criteria, which means the primal and dual residuals are reduced at a ideal small value, respectively, as
%\begin{equation}
%\begin{aligned}
%\|\W^{t} - \Z^{t}\|_{2} \le \epsilon^{\textrm{pri}}\quad \textrm{and} \quad
%\|\Z^{t+1} - \Z^{t}\|_{2} \le \epsilon^{\textrm{dual}},
%\end{aligned}
%\label{eq12}
%\end{equation}
%where $\|\W^{t} - \Z^{t}\|$ denotes the primal residual and $\|\Z^{t+1} - \Z^{t}\|$ denotes the dual residual, $\epsilon^{\textrm{pri}} \geq 0$ and $\epsilon^{\textrm{dual}} \geq 0$ are feasibility tolerances for the primal and dual feasibility conditions, respectively. One way to design feasibility tolerances is using the method named absolute and relative criterion, which is combining the absolute tolerance and the relative tolerance together to compute the primal and dual target residuals. 

\subsection{Fine-Tuning}
After ADMM-regularized training, we first decompose the trained uncompressed DNN model into TT format. Here the decomposition is performed with the target TT-ranks $\bm{r}^*$ for tensor cores. Because the ADMM optimization procedure has already imposed the desired low TT-rank structure to the uncompressed model, such direction decomposition, unlike their counterpart in the existing TT-format DNN training, will not bring significant approximation error (More details will be analyzed in Section \ref{subsec:converge}). Then, the decomposed TT-format model is fine-tuned using standard SGD. Notice that in the fine-tuning phase the loss function is  $\ell(\{\bm{\mathcal{G}}_i\})$ without other regularization term introduced by ADMM. Typically this fine-tuning phase is very fast with requiring only a few iterations. This is because the decomposed TT model at the starting point of this phase already has very closed accuracy to the original uncompressed model.

% Moreover, unlike some state-of-the-art methods such as \cite{li2020heuristic, garipov2016ultimate} which uses a well pretrained model to enhance performance, our framework does not need any pretrained model.

%-------------------------------------------------------------------------
\section{Experiments}
\label{sec:exp}

To demonstrate the effectiveness and generality of the proposed compression framework, we evaluate different DNN models in different computer vision tasks. For image classification tasks, we evaluate multiple CNN models on MNIST, CIFAR-10, CIFAR-100 and ImageNet datasets \cite{lecun1998gradient,krizhevsky2009learning,deng2009imagenet}. For video classification tasks, we evaluate different LSTM models on UCF11 and HMDB51 datasets \cite{liu2009recognizing,kuehne2011hmdb}. We follow the same rank selection scheme adopted in prior works -- set ranks to satisfy the need of the targeted compression ratio. To simplify selection procedure, most of the ranks in the same layer are set to equal.

\subsection{Convergence and Sensitivity Analysis}

\label{subsec:converge}

\begin{figure*}[t]
     \centering
     \begin{subfigure}[b]{0.33\textwidth}
         \centering
         \vskip -2mm
         \includegraphics[width=\linewidth]{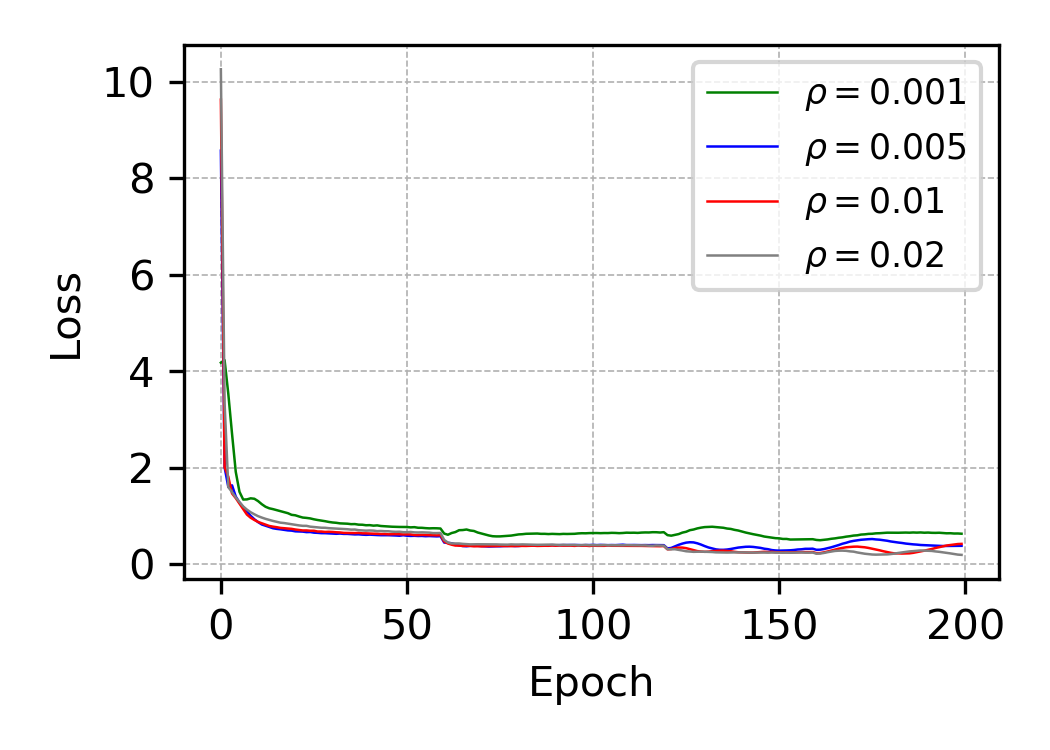}
         \vskip -4mm
         \caption{Training loss.}
         \label{fig:admm_curves1}
     \end{subfigure}
     \hfill
     \begin{subfigure}[b]{0.33\textwidth}
         \centering
         \vskip -2mm
         \includegraphics[width=\linewidth]{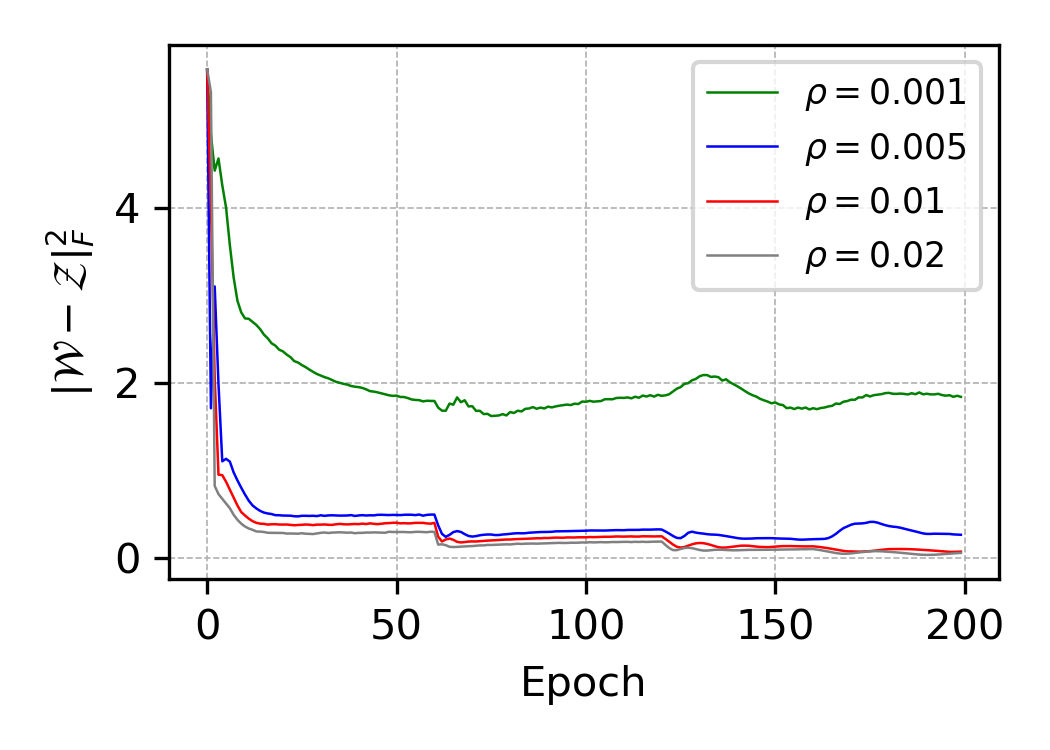}
         \vskip -4mm
         \caption{$\|\W-\bm{\mathcal{Z}}\|_F^2$.}
         \label{fig:admm_curves2}
     \end{subfigure}
     \hfill
     \begin{subfigure}[b]{0.33\textwidth}
         \centering
         \vskip -2mm
         \includegraphics[width=\linewidth]{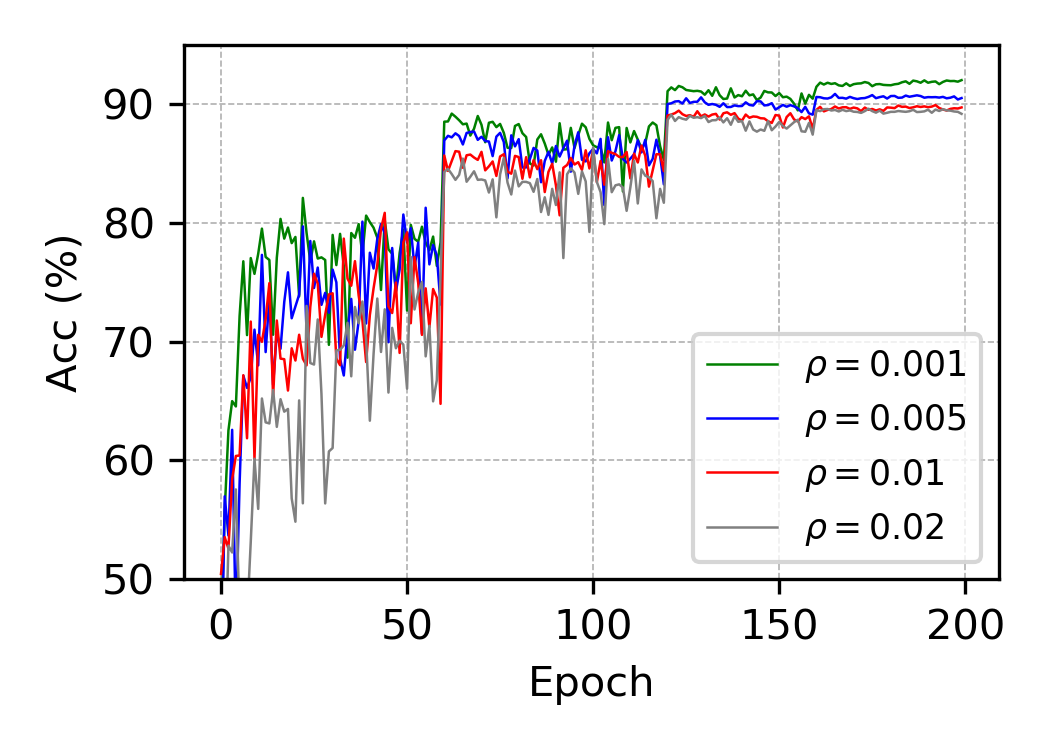}
         \vskip -4mm
         \caption{Top-1 test accuracy.}
         \label{fig:admm_curves3}
     \end{subfigure}
     \vskip -3mm
        \caption{Training loss, Frobenius norm and test accuracy in ADMM-regularized training procedure with different $\rho$. }
        \label{fig:admm_curves}
    \vspace{-4mm}
\end{figure*}

As shown in (\ref{eqn:lagran}), $\rho$ is the additional hyperparameter introduced in the ADMM-regularized training phase. To study the effect of $\rho$ on the performance as well as facilitating hyperparameter selection, we study the convergence and sensitivity of the ADMM-regularized training for ResNet-32 model with different $\rho$ settings on CIFAR10 dataset.

\textbf{Convergence.} Figure \ref{fig:admm_curves1} shows the loss curves in the ADMM-regularized training phase. It is seen that different curves with very different $\rho$ values (e.g. 0.001 vs 0.02), exhibit very similar convergence speed. This phenomenon therefore demonstrates that $\rho$ has little impact on the convergence of ADMM-regularized training.

\textbf{Sensitivity.} Considering the similar convergence behavior does not necessarily mean that different $\rho$ would bring the similar accuracy, we then analyze the performance sensitivity of ADMM-regularized training with respect to $\rho$. Notice that ideally after ADMM-regularized training, $\W$, though in the uncompressed format, should exhibit strong low TT-rank characteristics and meanwhile enjoy high accuracy. Once $\W$ meets such two criteria simultaneously, that means TT-cores $\{\bm{\mathcal{G}}_i\}$, whose initialization is decomposed from $\W$, will already have  high accuracy even before fine-tuning.

To examine the required low TT-rank behavior of $\W$, we observe $\|\W-\Z\|_F^2$, which measures the similarity between $\W$ and $\Z$,  in the ADMM-regularized training (see Figure \ref{fig:admm_curves2}). Since according to (\ref{eqn:z_update}) $\Z$ is always updated with low TT-rank constraints, the curves shown in Figure \ref{fig:admm_curves2} reveal that $\W$ indeed quickly exhibits low TT-rank characteristics during the training, except when $\rho=\textrm{0.001}$. This phenomenon implies that to ensure the weight tensors are well regularized to the target TT-ranks by ADMM,  $\rho$ should not be too small (e.g. less than 0.001). 
On the other hand, Figure \ref{fig:admm_curves3} shows the test accuracy of $\W$ as training progresses. Here it is seen that smaller $\rho$ tends to bring better performance. Based on these observations, $\rho=\textrm{0.005}$ can be an appropriate choice to let the trained $\W$ meet the aforementioned two criteria.

\begin{table}[t]
\begin{center}
\begin{tabular}{|l|c|c|c|c|}
\hline
\multicolumn{1}{|c|}{\multirow{2}{*}{\textbf{Model}}} & \multirow{2}{*}{\hspace{-1mm}\makecell{\textbf{Comp.}\\\textbf{Method}}\hspace{-1mm}} & \multirow{2}{*}{\makecell{\textbf{Top-1}\\\textbf{(\%)}}}  &  \multirow{2}{*}{\makecell{\textbf{Comp.}\\\textbf{Ratio}}}\\ 
& & & \\
\hline\hline
Uncompressed & - & 99.21  & 1.0$\times$ \\
\hline
Standard TR\cite{wang2018wide}   & \multirow{3}{*}{TR}  & 99.10   & 10.5$\times$  \\
PSTRN-M\cite{li2020heuristic} &    & 99.43  &  16.5$\times$        \\
PSTRN-S\cite{li2020heuristic} &    & 99.51  &  6.5$\times$        \\
\hline
Standard TT\cite{garipov2016ultimate}& \multirow{3}{*}{TT}  & 99.07 & 17.9$\times$ \\
 \textbf{Ours} & & 99.48 & \textbf{17.9}$\times$      \\ 
  \textbf{Ours} & & \textbf{99.51} &  8.3$\times$      \\
 \hline
\end{tabular}
\end{center}
\vspace{-6mm}
\caption{LeNet-5 on MNIST dataset using different TT/TR-format compression approaches.}
\label{tbl:mnist}
\vspace{-2mm}
\end{table}

%Additionally, we plot the loss curves of TT models trained by three methods in Figure XX. The first is to randomly initialize the TT model then directly train, which is the most common. The second method initializes the TT model by decomposing a pretrained dense model to TT-format, then retrain it. The third one is using our compressing framework to initialize the TT model from ADMM regularization. By the curves, we can see that the convergence speed using the proposed framework is much faster that the others, and eventually the loss value is also smaller than the other two.

\subsection{Image Classification}
\textbf{MNIST.} Table \ref{tbl:mnist} shows the experimental results of LeNet-5 model \cite{lecun1998gradient} on MNIST dataset. We compare our ADMM-based TT-format model with the uncompressed model as well as the state-of-the-art TT/TR-format works. It is seen that our ADMM-based compression can achieve the highest compression ratio and the best accuracy.

\begin{table}[t]
\begin{center}
\begin{tabular}{|l|c|c|c|}
\hline
 \multicolumn{1}{|c|}{\multirow{2}{*}{\textbf{Model}}} & \multirow{2}{*}{\makecell{\textbf{Comp.}\\\textbf{Method}}} & \multirow{2}{*}{\makecell{\textbf{Top-1}\\\textbf{(\%)}}}  &  \multirow{2}{*}{\makecell{\textbf{Comp.}\\\textbf{Ratio}}}\\
& & & \\
 \hline\hline
 \multicolumn{4}{|c|}{ResNet-20} \\  \hline
 Uncompressed & - & 91.25 & 1.0$\times$\\
\hline
Standard TR\cite{wang2018wide, li2020heuristic}& \multirow{4}{*}{TR} & 87.5    & 5.4$\times$      \\
TR-RL\cite{cheng2020novel}&    & 88.3   & 6.8$\times$     \\
PSTRN-M\cite{li2020heuristic}  &    & 88.50    & 6.8$\times$    \\
PSTRN-S\cite{li2020heuristic} &      & 90.80       & 2.5$\times$   \\
\hline
Standard TT\cite{garipov2016ultimate}& \multirow{3}{*}{TT}  & 86.7  & 5.4$\times$        \\ 
\textbf{Ours} &   & 91.03  & \textbf{6.8}$\times$      \\
\textbf{Ours} &   & \textbf{91.47} & 4.5$\times$ \\ \hline\hline
 \multicolumn{4}{|c|}{ResNet-32} \\  \hline
Uncompressed & - & 92.49 &  1.0$\times$\\
\hline
Standard TR\cite{wang2018wide}& \multirow{3}{*}{TR} & 90.6  & 5.1$\times$        \\
PSTRN-M\cite{li2020heuristic}  &     & 90.6         & 5.8$\times$   \\
PSTRN-S\cite{li2020heuristic} &      & 91.44            & 2.7$\times$   \\
\hline
Standard TT\cite{garipov2016ultimate,wang2018wide}& \multirow{3}{*}{TT} & 88.3   & 4.8$\times$    \\
\textbf{Ours}&    & 91.96 & \textbf{5.8}$\times$ \\
\textbf{Ours} &    & \textbf{92.87} & 4.8$\times$ \\
 \hline
\end{tabular}
\end{center}
\vspace{-6mm}
\caption{ResNet-20 and ResNet-32 on CIFAR-10 dataset using different TT/TR-format compression approaches.}
\footnotesize{}
\label{tbl:cifar10}
\vspace{-4mm}
\end{table}

\textbf{CIFAR-10.} Table \ref{tbl:cifar10} compares our ADMM-based TT-format ResNet-20 and ResNet-32 models with the state-of-the-art TT/TR-format works on CIFAR-10 dataset. For ResNet-20, it is seen that standard training on TT/TR-format models causes severe accuracy loss. Even for the state-of-the-art design using some advanced techniques, such as heuristic rank selection (PSTRN-M/S) and reinforcement learning (TR-RL), the performance degradation is still huge, especially with high compression ratio 6.8$\times$. On the other side, with the same high compression ratio our ADMM-based TT-format model has only 0.22\% accuracy drop, which means 2.53\% higher than the state-of-the-art PSTRN-M. Furthermore, with moderate compression ratio 4.5$\times$ our method can even outperform the uncompressed model with 0.22\% accuracy increase.

For ResNet-32, again, standard training on compressed models using TT or TR decomposition causes huge performance degradation. The state-of-the-art PSTRN-S/M indeed brings performance improvement, but the test accuracy is still not satisfied. Instead, our highly compressed (5.8$\times$) TT-format model only has 0.53\% accuracy loss, which means it has 1.36\% higher accuracy than PSTRN-M with the same compression ratio. More importantly, when compression ratio is relaxed to 4.8$\times$, our ADMM-based TT-format model achieves 92.87\%, which is even 0.38\% higher than the uncompressed model.

\begin{table}[t]
\begin{center}
\begin{tabular}{|l|c|c|c|}
\hline
 \multicolumn{1}{|c|}{\multirow{2}{*}{\textbf{Model}}} & \multirow{2}{*}{\makecell{\textbf{Comp.}\\\textbf{Method}}} & \multirow{2}{*}{\makecell{\textbf{Top-1}\\\textbf{(\%)}}} & \multirow{2}{*}{\makecell{\textbf{Comp.}\\\textbf{Ratio}}}\\
& & &\\
 \hline\hline
 \multicolumn{4}{|c|}{ResNet-20} \\  \hline
 Uncompressed & - & 65.4 & 1.0$\times$ \\
 \hline
Standard TR\cite{wang2018wide,li2020heuristic} &  \multirow{3}{*}{TR} & 63.55   & 4.7$\times$       \\ 
  PSTRN-M\cite{li2020heuristic}  &     & 63.62  & 4.7$\times$ \\  
  PSTRN-S\cite{li2020heuristic}  &     & 66.13  & 2.3$\times$  \\ 
  \hline
  Standard TT\cite{garipov2016ultimate} & \multirow{3}{*}{TT}   & 61.64   & 5.6$\times$        \\ 
  \textbf{Ours} &  & 64.92 & \textbf{5.6}$\times$ \\ 
  \textbf{Ours} &  & \textbf{67.36}& 2.3$\times$ \\ \hline\hline
\multicolumn{4}{|c|}{ResNet-32} \\  \hline
Uncompressed & - & 68.10  & 1$\times$\\
\hline
Standard TR\cite{wang2018wide} & \multirow{3}{*}{TR} & 66.70 & 4.8$\times$       \\  
PSTRN-M\cite{li2020heuristic} &   & 66.77   & 5.2$\times$       \\  
PSTRN-S\cite{li2020heuristic} &   & 68.05   & 2.4$\times$        \\ 
\hline
Standard TT\cite{garipov2016ultimate,wang2018wide} & \multirow{3}{*}{TT} & 62.90  & 4.6$\times$ \\ 
\textbf{Ours} &  & 67.17 &\textbf{5.2}$\times$ \\
\textbf{Ours} &  & \textbf{70.31}  & 2.4$\times$      \\   \hline
\end{tabular}
\end{center}
\vskip -6mm
\caption{ResNet-20 and ResNet-32 on CIFAR-100 dataset using different TT/TR-format compression approaches.}
\label{tbl:cifar100}
\vspace{-4mm}
\end{table}

\textbf{CIFAR-100.} Table \ref{tbl:cifar100} shows the experimental results on CIFAR-100 dataset. Again, our ADMM-based TT-format model outperforms the state-of-the-art work. For ResNet-20, with even higher compression ratio (Our 5.6$\times$ vs 4.7$\times$ in PSTRN-M), our model achieves 1.3\% accuracy increase. With 2.3$\times$ compression ratio, our model achieves 67.36\% Top-1 accuracy, which is even 1.96\% higher than the uncompressed model. For ResNet-32, with the same 5.2$\times$ compression ratio, our approach brings 0.4\% accuracy increase over the state-of-the-art PSTRN-M. With the same 2.4$\times$ compression ratio, our approach has 2.26\% higher accuracy than PSTRN-S. Our model even outperforms the uncompressed model with 2.21\% accuracy increase.

\begin{table}[t]
\begin{center}
\begin{tabular}{|l|c|c|c|}
\hline
 \multicolumn{1}{|c|}{\multirow{2}{*}{\textbf{Model}}} & \multirow{2}{*}{\makecell{\textbf{Comp.}\\\textbf{Method}}} & \multirow{2}{*}{\makecell{\textbf{Top-5}\\\textbf{(\%)}}} &  \multirow{2}{*}{\textbf{FLOPs$\downarrow$}}\\
& & & \\ \hline\hline
\multicolumn{4}{|c|}{ResNet-18} \\  \hline
 Uncompressed & - & 89.08  & 1.00$\times$ \\
\hline
Standard TR\cite{wang2018wide} &  \multirow{1}{*}{TR}  & 86.29  & 4.28$\times$\\ 
\hline
TRP\cite{xu2020trp}& \multirow{2}{*}{\makecell{Matrix\\SVD}} & 86.74 &  2.60$\times$ \\
TRP+Nu\cite{xu2020trp}& & 86.61 & 3.18$\times$\\
\hline
DACP\cite{zhuang2018discrimination} & \multirow{4}{*}{Pruning} & 87.60 &  1.89$\times$\\
%SFP\cite{he2018soft}&  & 87.78 & - & 1.72$\times$\\
FBS\cite{gao2019dynamic} & & 88.22 &  1.98$\times$ \\
FPGM\cite{he2019filter} & & 88.53 &  1.72$\times$\\
DSA\cite{ning2020dsa} & & 88.35 &  1.72$\times$\\
\hline
Standard TT\cite{garipov2016ultimate} & \multirow{3}{*}{TT}  &   85.64  & 4.62$\times$        \\  
\textbf{Ours} &  & 87.47 & \textbf{4.62}$\times$ \\ 
    \textbf{Ours} &  & \textbf{89.08} & 2.47$\times$ \\
\hline
\end{tabular}
\end{center}
    \vspace{-6mm}
\caption{ResNet-18 on ImageNet dataset using compression approaches. We do not list PSTRN-M/S since \cite{li2020heuristic} does not report results on ImageNet. Also the listed pruning and SVD works do not report compression ratios in their papers. The uncompressed baseline model is from Torchvision. Note that the reported Top-5 accuracy of \cite{gao2019dynamic,he2019filter} in this table are obtained from pruning the baselines with higher accuracy.}
\label{tbl:ImageNet}
\vspace{-5mm}
\end{table}

\textbf{ImageNet.} Table \ref{tbl:ImageNet} shows the results of compressing ResNet-18 on ImageNet dataset. Because no prior TT/TR compression works report results on this dataset, we use standard TT and TR-based training in \cite{wang2018wide,garipov2016ultimate} for comparison. We also compare our approach with other compression methods, including pruning and matrix SVD. Since these works report FLOPs reduction instead of compression ratio, we also report FLOPs reduction brought by tensor decomposition. It is seen that with the similar FLOPs reduction ratio (4.62$\times$), our ADMM-based TT-format model has 1.83\% and 1.18\% higher accuracy than standard TT and TR, respectively. Compared with other compression approaches with non-negligible accuracy loss, our ADMM-based TT-format models achieve much better accuracy with more FLOPs reduction. In particular, with 2.47$\times$ FLOPs reduction, our model has the same accuracy as the uncompressed baseline model.

\subsection{Video Recognition}

\textbf{UCF11.} In this experiment, we use the same uncompressed LSTM model, data pre-processing and experimental settings adopted in \cite{ye2018learning,pan2019compressing}. To be consistent with \cite{ye2018learning,pan2019compressing}, only the ultra-large input-to-hidden layer is compressed for fair comparison. Table \ref{tbl:UCF11} compares our ADMM-based TT-format LSTM with the uncompressed model and the existing TT-LSTM \cite{yang2017tensor} and TR-LSTM \cite{pan2019compressing}. Note that \cite{li2020heuristic} does not report the performance of PSTRN-M/S on UCF11 dataset.

From Table \ref{tbl:UCF11} , it is seen that both TT-LSTM and TR-LSTM provide remarkable performance improvement and excellent compression ratio. As analyzed in \cite{yang2017tensor}, such huge improvement over the uncompressed model mainly comes from the excellent feature extraction capability of TT/TR-format LSTM models on the ultra-high-dimensional inputs. Compared with these existing works, our ADMM-based TT-format model achieves even better performance. With fewer parameters, our method brings 2.1\% higher top-1 accuracy than the state-of-the-art TR-LSTM.

\begin{table}[t]
\begin{center}
\begin{tabular}{|l|c|c|c|c|}
\hline
 \multicolumn{1}{|c|}{\multirow{2}{*}{\textbf{Model}}} & \multirow{2}{*}{\hspace{-1.5mm}\makecell{\textbf{Comp.}\\\textbf{Method}}\hspace{-1.5mm}} & \multirow{2}{*}{\makecell{\textbf{Top-1}\\\textbf{(\%)}}} & \multirow{2}{*}{\hspace{-1mm}\textbf{\# Para.}\hspace{-1mm}} &  \multirow{2}{*}{\hspace{-1mm}\makecell{\textbf{Comp.}\\\textbf{Ratio}}\hspace{-1mm}}\\
& & & &\\ \hline\hline
Uncompressed & - &  69.7 & 59M & 1.0$\times$ \\
\hline
TR-LSTM\cite{pan2019compressing} & \multirow{1}{*}{TR}  & 86.9          & 1,725   &34.2K$\times$\\ 
\hline
TT-LSTM\cite{yang2017tensor} & \multirow{2}{*}{TT}  & 79.6 & 3,360  & 17.6K$\times$ \\ 
 \textbf{Ours} &  & \textbf{89.0} & \textbf{1,656} &\textbf{35.6K}$\times$ \\  \hline
 
\end{tabular}
\end{center}
\vspace{-6mm}
\caption{LSTM on UCF11 dataset using different TT/TR-format compression approaches.}
%\cite{li2020heuristic} does not report performance of PSTRN-M/S.}
\label{tbl:UCF11}
\vspace{-4mm}
\end{table}

\textbf{HMDB51.} To be consistent with the setting adopted in \cite{li2020heuristic,pan2019compressing}, in this experiment we use the same Inception-V3 as the front-end pre-trained CNN model, and the same back-end uncompressed LSTM model. For fair comparison, we follow the compression strategy adopted in \cite{li2020heuristic,pan2019compressing} as only compressing the ultra-large input-to-hidden layer of LSTM.

Table \ref{tbl:HMDB51} summarizes the experimental results. It is seen that comparing with the state-of-the-art TT/TR-format designs, our ADMM-based TT-format model shows excellent performance. With the highest compression ratio (84.0$\times$), our model achieves 64.09\% top-1 accuracy. Compared with the state-of-the-art TR-LSTM, our model brings 3.35$\times$ more compression ratio with additional 0.29\% accuracy increase.

\begin{table}[t]
\begin{center}
\begin{tabular}{|l|c|c|c|c|}
\hline
 \multicolumn{1}{|c|}{\multirow{2}{*}{\textbf{Model}}} & \multirow{2}{*}{\hspace{-1mm}\makecell{\textbf{Comp.}\\\textbf{Method}}\hspace{-1mm}} & \multirow{2}{*}{\makecell{\textbf{Top-1}\\\textbf{(\%)}}} & \multirow{2}{*}{\textbf{\# Para.}} &  \multirow{2}{*}{\hspace{-1mm}\makecell{\textbf{Comp.}\\\textbf{Ratio}}\hspace{-1mm}}\\
& & & &\\ \hline\hline
Uncompressed & - & 62.9 & 16.8M & 1.0$\times$ \\
\hline
TR-LSTM\cite{pan2019compressing} & \multirow{3}{*}{TR}   &   63.8      &  0.67M   & 25.0$\times$        \\  
PSTRN-M\cite{li2020heuristic}  &  &   59.67   & 0.36M   & 46.7$\times$       \\ 
PSTRN-S\cite{li2020heuristic} &  & 60.04  &    0.48M  &  34.7$\times$ \\
\hline
TT-LSTM\cite{yang2017tensor} &\multirow{2}{*}{TT}   &  62.24   &  0.67M   & 25.0$\times$        \\ 
  \textbf{Ours} &  & \textbf{64.09} & \textbf{0.20M} & \textbf{84.0}$\times$ \\ \hline
\end{tabular}

\end{center}
\vspace{-6mm}
\caption{LSTM on HMDB51 dataset using different TT/TR-format compression approaches. }
\label{tbl:HMDB51}
\vspace{-3mm}
\end{table}

\subsection{Discussion on Tensor Format and Generality}
% \textbf{Target Rank.} The selection of target tensor rank is a fundamental problem in tensor decomposition-based compression methods even low-rank matrix decomposition-based methods, which is merely touched by prior works. Our method aims to minimize performance degradation with fixed tensor ranks. Hence, in experiments we select the target tensor ranks according to the prior works such as [][][], which already obtain a sort of relatively optimal tensor ranks.

\textbf{Why Choosing TT-format.} Recently several state-of-the-art tensor decomposition-based compression works \cite{wang2018wide,pan2019compressing,ye2018learning} report that TT decomposition is inferior to other advanced approach (e.g. TR) on DNN compression, in terms of compression ratio and test accuracy. To fully demonstrate the excellent effectiveness of our approach, in this paper we choose TT, the tensor format that is believed to be not the best for model compression, and adapt the ADMM-regularized compression framework to TT-format. As presented in the experimental results, all the ADMM-based TT-format models consistently outperform the existing TT/TR-format models with higher accuracy and higher compression ratio over different datasets, thereby comprehensively demonstrating the huge benefits brought by our proposed framework. 

\textbf{Generality of Our Framework.} Although in this paper our focus is to compress TT-format DNN models, because ADMM is a general optimization technique, \ul{our proposed framework is very general and can be easily applied for model compression using other tensor decomposition approaches, such as Tensor Ring (TR), Block-term (BT), Tucker etc}. To adapt to other tensor decomposition scenario, the main modification on our proposed framework is to modify the Euclidean projection (Algorithm \ref{alg:ttsvd}) to make the truncating methods being compatible to the corresponding tensor decomposition methods.

%\subsection{ADMM}
%The formulated optimization problem is extremely difficult to be optimized since it is nonconvex and non-differentiable. Fortunately, ADMM provides an alternative solution with better convergence property e.g. faster convergence or convergence to a point with better objective value \cite{boyd2011distributed}, compared with other local optimization methods. In Figure \ref{fig:admm_curves1}, our experiment show that ADMM reaches global optimization with variant penalty parameters. Additionally, a recent pruning work \cite{zhang2018systematic} also prove that ADMM works very well for such nonconvex problem.

\section{Conclusion}
In this paper, we present a systematic compression framework for tensor-format DNNs using ADMM. Under the framework, the tensor decomposition-based DNN model compression is formulated to a nonconvex optimization problem with constraints on target tensor ranks. By performing ADMM to solve this problem,  a uncompressed but low tensor-rank model can be obtained, thereby finally bringing the decomposed high-accuracy TT-format model. Extensive experiments for image and video classification show that our ADMM-based TT-format models consistently outperform the state-of-the-art works in terms of compression ratio and test accuracy.

\section*{Acknowledgements}
This work was partially supported by National Science Foundation under Grant CCF-1955909.

\clearpage
{\small
\bibliographystyle{ieee_fullname}
\bibliography{egbib}
}

\end{document}